\documentclass[10pt,twocolumn,letterpaper]{article}

\usepackage{iccv}
\usepackage{times}
\usepackage{epsfig}
\usepackage{graphicx}
\usepackage{amsmath}
\usepackage{amssymb}
\usepackage{algorithm}
\usepackage{algorithmic}
\usepackage{bbm}
\usepackage{booktabs}
\usepackage{amsfonts}
\usepackage{mathrsfs}
\usepackage{multirow}
\usepackage{subfigure}
\DeclareUnicodeCharacter{2212}{-}

\usepackage[pagebackref=true,breaklinks=true,letterpaper=true,colorlinks,bookmarks=false]{hyperref}

\iccvfinalcopy 

\def\iccvPaperID{406} 

\begin{document}

\title{KFC: Kinship Verification with Fair Contrastive Loss and Multi-Task Learning}

\author{Jia Luo Peng,~ Keng Wei Chang,~ Shang-Hong Lai\\\vspace{-8pt}{\small~}\\
National Tsing Hua University~~~ \\
{\small{\tt{\{garynlfd, percyx987654321\}@gapp.nthu.edu.tw} \tt{lai@cs.nthu.edu.tw}}}
}

\maketitle
\thispagestyle{empty}

\begin{abstract}
Kinship verification is an emerging task in computer vision with several potential applications. However, there is a lack of large kinship datasets to train a discriminative and robust model, which is a major limitation for this problem. Moreover, face verification is known to exhibit bias in skin colors and ethics, which has not been fully resolved by previous works.  In this paper, we propose a multi-task learning model structure with attention module to improve the accuracy, which surpasses state-of-the-art performance. In addition, our fairness-aware contrastive loss function combined with a debias term and adversarial learning greatly mitigates racial bias, thus significantly improving the fairness. In the experiment, we build a large dataset by combining several existing kinship datasets. Exhaustive experimental evaluation demonstrates the effectiveness and superior performance of the proposed KFC in both standard deviation and accuracy at the same time. Code is available at \url{https://github.com/garynlfd/KFC}
\end{abstract}

\section{Introduction}

\label{sec:intro}
Facial kinship verification\cite{10.1007/s11263-022-01605-9}\cite{robinson20215th} has been an emerging research area, which can be used for missing children, forensic analysis, multimedia analysis, and personal photos app management (e.g., creating a family tree). Human faces contain not only family information but also racial traits, but racial traits often get ignored while researchers usually focus more on accuracy performance. However, such unfairness has been an infamous problem in AI systems and involves in many social problems in recent years, e.g., healthcare system\cite{2019}, hiring algorithm\cite{2018}, recidivism judgement\cite{2016}, etc. 

Many previous works have developed methods for improving fairness in face recognition and face verification. For example, \cite{park2022fair}\cite{xu2021consistent} proposed fairness-aware loss functions. The former exploits the interrelation between anchor and sample to design a sensitive attribute removing loss function, while the latter uses instance FPR in loss function to constrain bias. \cite{wang2022fairnessaware} implemented post-processing data perturbation without changing their parameters and structures that can
hide the information of protected attributes. \cite{gong2020mitigating} proposed to include demographic-adaptive layers that make the model generate face representations for every demographic group. 

\begin{figure}[h]
\centering
\includegraphics[scale=0.15]{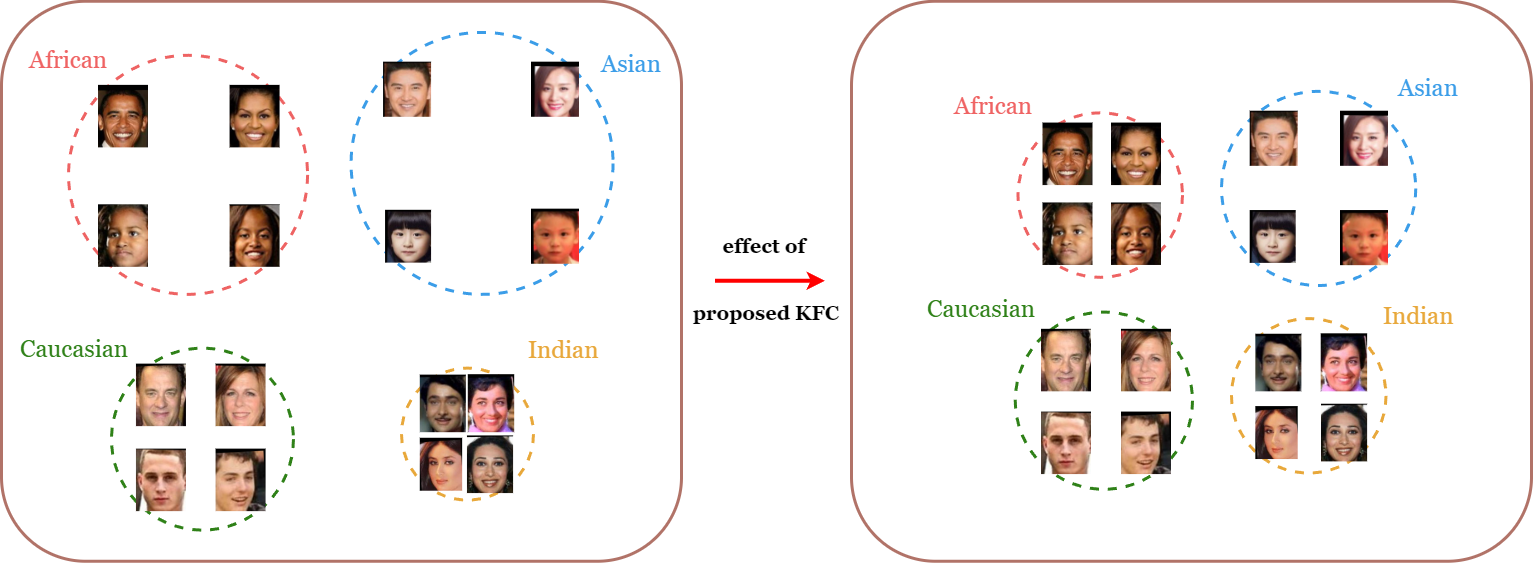}
\caption{The schematic diagram for improving the fairness in kinship verification. Our method can effectively adjust the intra-class compactness and inter-class discrepancy in the feature space. We mitigate racial bias by balancing four races' intra-class and inter-class angle and making them as consistent as possible.}
\label{fig:schematic diagram}
\end{figure}

Kinship verification is a subfield of face verification that involves identifying the degree of relatedness between individuals based on their unique facial features. Siamese neural network\cite{1467314}\cite{Chicco2021}\cite{6909616}\cite{10.5555/2987189.2987282} is an intuitive network that uses weight-sharing structure to extract features, which is an effective kinship verification network that focuses on similar facial traits. Attention scheme\cite{inproceedings}\cite{su2023kinship}\cite{YAN2019169} is another popular method in kinship verification. Since facial features are crucial for determining the relationship between two face images, it is important to make the model focus on the most critical regions.

Based on the above idea, we propose an approach to improve racial fairness while  achieving higher accuracy in kinship verification. Furthermore, to deal with fairness issues and small dataset constraint, we combine multiple kinship datasets and label every individual's race, constructing a large kinship dataset with race label, called KinRace dataset. To boost accuracy and fairness at the same time, we propose a fairness-aware loss function in a multi-task learning framework. To improve the kinship verification accuracy, we use attention module that makes model focus on the most representative facial regions for feature representation learning. For fairness, we reverse the gradient of race classification branch to remove the racial information in the feature vector and design a fairness-aware contrastive loss function which can mitigate pairwise bias and significantly decrease the standard deviation in four races. 

With the above feature learning schemes, our method successfully finds a good balance for the distribution of the intra-class and inter-class distances for all the races. As shown in \autoref{fig:schematic diagram}, the proposed feature learning approach effectively improves racial fairness and equilibrates the compactness for all races. Compared with the previous works, our method firstly has an innovative model structure to tackle the kinship verification task. Secondly, it utilizes two different debias techniques to improve the fairness. Finally, all these methods are integrated to enhance both fairness and accuracy performance on our integrated dataset.

In summary, the main contributions of this paper are 
three-fold:
\begin{itemize}
    \item To the best of our knowledge, this is the first work proposed to mitigate bias and achieve state-of-the-art accuracy simultaneously for kinship verification.
    \item Our fairness-aware contrastive loss function automatically mitigates the pairwise bias and balances the degree of compactness of every race, which improves racial fairness.
    \item We build a large kinship dataset with racial labels from several public kinship datasets, which is targeted for the research on the fairness of kinship verification.
\end{itemize}

\section{Related Work}\label{c:related}

\subsection{Kinship Verification}

Kinship verification has attracted a number of researchers  in the past few years, due to the emergence of datasets\cite{Buffalo_TMM_Kinship}\cite{Siyu_IJCAI11_Kinship}\cite{Ming_CVPR11_Genealogical}\cite{6247978}\cite{6738614}\cite{robinson2016families}, challenges\cite{robinson20215th}, and the potential in real-world applications. Traditional methods used hand-crafted features to extract local texture information. \cite{inproceedingsLBP} used binary code to extract local binary patterns that consider both shape and texture information, which are invariant to grey scale transformation. \cite{inproceedings2} used two texture descriptors Local Directional Pattern (LDP) and Local Phase Quantization(LPQ) for feature extraction. By capturing different aspects of texture information and characterizing various patterns and structures present in images, they produces better performance than the previous approaches. Shallow metric learning method\cite{8019326} selected critical feature patches and discards trivial ones by incorporating AdaBoost and SVM, and each weak classifier of AdaBoost linearly combines as a strong classifier to select various features and create a discriminative model. \cite{5652590} selected hand-crafted features such as colors and distances between the five senses, focusing on those features they deemed discriminative. This approach can be regarded as a simple and shallow way to extract features before deep learning.

Deep learning methods have revolutionized various fields by automatically extracting highly effective features, leading to significant improvements in accuracy. Building upon this foundation, researchers have proposed innovative techniques to leverage deep learning for kinship verification tasks. \cite{yu2020deep} proposed a feature fusion method that uses discriminative features from backbone network and fuses the features to determine if two face images are with kinship relationship or not. \cite{9666944} adopted ArcFace\cite{Deng_2022} as the backbone model pre-trained on MS-Celeb-1M\cite{guo2016msceleb1m} to obtain more representative features. Moreover, they used supervised contrastive loss function to contrast samples against each other and hyperparameter to focus on hard sample, thus enhancing the ability to distinguish the kinship relation. \cite{su2023kinship} successfully enhance the accuracy of kinship verification task by leveraging attention mechanism. They combined attention mechanism with backbone to focus on the most discrminative part(e.g., five senses) of facial image. They also proposed a new loss function that combined contrastive loss and the attention map they created from the attention mechanism.


\subsection{Bias Mitigation}

The ongoing effort to tackle racial bias within AI systems has prompted researchers to explore a range of innovative approaches. These include developing new algorithms, modifying the structures of models, and proposing novel loss functions. One notable study \cite{raff2018gradient}, devised fair features using an adversarial learning technique. This method involved the incorporation of a gradient reversal layer, effectively flipping the gradient of the classification head for sensitive attributes. This strategic move encouraged the model's encoder to generate features devoid of sensitive information, thus reducing potential bias. Similarly, another research \cite{gong2020jointly} leveraged adversarial learning to attain discriminative feature representation, simultaneously disentangling features into four distinct attributes. This process of disentanglement aimed to preserve crucial attributes while discarding unfair ones. By carefully manipulating the feature space, the model could successfully eliminate biases linked with sensitive attributes. \cite{wang2022fairnessaware} introduced an approach with the aim of mitigating bias in deployed models. Unlike previous state-of-the-art methods that focused on altering the deployed models, they took a different route by concentrating on perturbing inputs. They employed a discriminator trained to differentiate fairness-related attributes from latent representations within the deployed models. Simultaneously, an adversarially trained generator worked to deceive the discriminator, ultimately generating perturbations that can conceal the information associated with protected attributes.

In addition to the use of adversarial learning, \cite{gong2020mitigating} proposed the incorporation of adaptive layers within the model structure. The introduced adaptive layer aimed to enhance representation robustness for different demographic groups. An automation module was integrated to determine the optimal usage of adaptive layers in various model layers, dynamically adjusting the network's behavior to cater to the unique requirements of different groups. This adaptability significantly contributed to enhancing both racial fairness and overall performance across diverse demographic groups.

Another approach, \cite{xu2021consistent} involved the modification of the softmax loss function with a novel penalty term to mitigate bias while concurrently improving accuracy. They achieved this by utilizing instance False Positive Rate as a surrogate for demographic False Positive Rate, eliminating the need for explicit demographic group labels.

Moreover, \cite{wang2022mixfairface}, shifted its focus from demographic group bias to identity bias. They combined the CosFace \cite{wang2018cosface} with bias difference to create a novel loss function. Their belief was that by targeting identity bias, they could solve the problem of skewed outcomes and treated all individuals impartially, striving for a comprehensive fairness that not dividing people based on their races. This innovative approach minimized identity bias without requiring sensitive attribute labels, thereby effectively enhancing fairness between demographic groups.

In conclusion, the pursuit of addressing racial bias within AI systems has spurred a multitude of creative strategies. These approaches encompass adversarial learning, the integration of adaptive layers, loss function modifications, and targeted bias reduction techniques, all of which collectively contribute to advancing fairness and equity in AI applications.

Drawing inspiration from the aforementioned studies, our research takes a step further by integrating two crucial domains: fairness and accuracy, with the goal of simultaneously improving both aspects. To achieve this objective, we introduce an attention mechanism and employ a multi-task model structure, while leveraging the power of adversarial learning and incorporating a fairness term in the loss function.

\section{Dataset Construction}\label{c:dataset construction}

\begin{figure*}[t]
\centering
\includegraphics[scale=0.35]{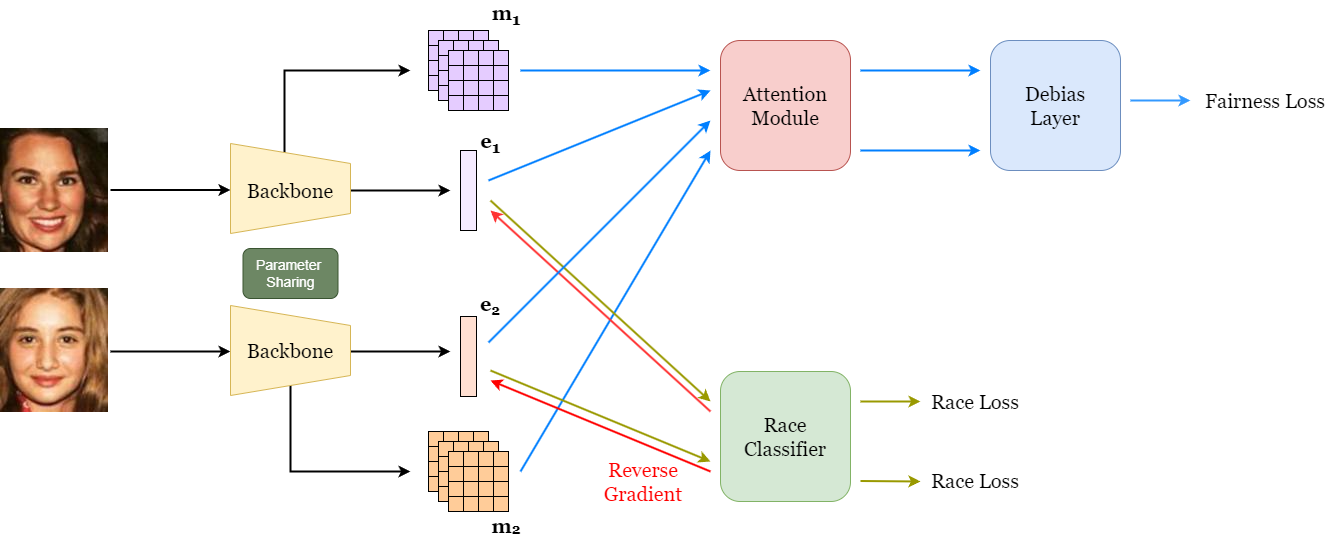}
\caption{Overview of the proposed KFC model structure. $e_1$ and $e_2$ are feature vectors from backbone, and they are used for race prediction with gradient reversal. $m_1$ and $m_2$ are feature maps computed from backbone. They are passed to attention module to extract important semantics and then passed to the debias layer to generate debias term in loss function.}
\label{fig:model structure}
\end{figure*}

We combine six kinship datasets to increase kinship verification dataset size and name the dataset KinRace. There are four small datasets: CornellKin\cite{5652590}, UBKinFace\cite{Buffalo_TMM_Kinship}\cite{Siyu_IJCAI11_Kinship}\cite{Ming_CVPR11_Genealogical}, KinFaceW-I\cite{6247978}, KinFaceW-II\cite{6247978}, and two large datasets: Family101\cite{6738614}, FIW\cite{robinson2016families}. While the data from KinFaceW can be easily classified using scores\cite{7393855}, we believe that the practical application of kinship verification will involve cropping images from the same photo. Additionally, the proportion of KinFaceW\cite{6247978},  data in our dataset is very small (0.62\%) , thus we argue that including KinFaceW\cite{6247978} in our combined dataset has negligible influence to our experimental results. Because most datasets only label four kinship types: Father-Son(FS), Father-Daughter(FD), Mother-Son(MS), and Mother-Daughter(MD), we only use these four kinship types in our mixed dataset. For Family101\cite{6738614}, some identities have too many images, so we limit the total number of images for each identity to at most 30. To deal with racial fairness, we manually label every identity's race. Following previous racial datasets\cite{Wang_2019_ICCV}, \cite{9512390}, we have four races in our dataset: African, Asian, Caucasian, and Indian. 

Other-race-effect (ORE) is a phenomenon that humans are better at recognizing faces of their race more readily than other races. Due to ORE, our labeling process was done by three different racial annotators (African, Asian, Caucasian). If two or three annotators label the same race to an identity, we take the values as ground truth. If all three annotators have different labeling results, we will not use this identity in our dataset. \autoref{tab:dataset race distribution} is the data distribution of our KinRace dataset. The total number of our image pairs is the largest among the existing kinship datasets, which means we can have more representative features and more generalizable experimental results. The race distribution is also similar to BUPT-Globalface \cite{9512390} whose racial distribution is approximately the same as the real distribution of the world’s population. For the correctness of experiments, our positive pairs (in the same family) must be the same race, which means we remove those mixed race samples.

Our motivation for constructing a mixed dataset arises from the limitations of large data availability and the absence of race labels. The reason we use only four races in our dataset is because some popular face datasets for fairness study, like RFW\cite{Wang_2019_ICCV} and BUPT\cite{9512390} datasets, use four races. Since we want to enable subsequent researchers to conduct studies on the same benchmark, we also use four races in our dataset.

Despite the BUPT dataset\cite{9512390} has a balanced representation of four races, biases still exist in the trained model, as shown in their paper (Table 7). Moreover, kinship datasets are not easily accessible, making it unavoidable to use an unbalanced dataset to simulate a challenging real-world setting. Following the approach of the BUPT dataset\cite{9512390}, we labeled the four races and aimed to reduce bias in kinship verification without relying on balanced data. The number of images per person is not well balanced. Since we aim to create a representative dataset, we intend to enlarge its size. However, we still control the number of images per person, limiting it to 30 to avoid overfitting.

The image quality of these six datasets we used are not the same, but the two main datasets have similar resolutions: Family101(120x150), FIW(108x124).
However, Table 5 in FaCoRNet\cite{su2023kinship} showed that there is no significant improvement in Kinship verification when considering data quality alone. Since image quality is crucial for face verification, we will explore more along this direction in our future work.

\begin{table}[h]
    \centering
    \scalebox{0.7}{
    \begin{tabular}{c|cccc|c}
         \toprule
         \textbf{ } & \textbf{African} & \textbf{Asian} & \textbf{Caucasian} & \textbf{Indian} & \textbf{sum} \\
         \midrule
         CornellKin\cite{5652590} & 8 & 56 & 72 & 3 & 139 \\
         UBKinFace\cite{Buffalo_TMM_Kinship}\cite{Siyu_IJCAI11_Kinship}\cite{Ming_CVPR11_Genealogical} & 18 & 192 & 173 & 0 & 383 \\
         KinFaceW-I\cite{6247978} & 19 & 327 & 172 & 4 & 522 \\
         KinFaceW-II\cite{6247978} & 55 & 96 & 788 & 35 & 974 \\
         Family101\cite{6738614} & 5554 & 6540 & 82820 & 25374 & 120288 \\
         FIW(train)\cite{robinson2016families} & 8353 & 3841 & 59028 & 681 & 71903 \\
         FIW(val)\cite{robinson2016families} & 2087 & 1398 & 30147 & 0 & 33632 \\
         FIW(test)\cite{robinson2016families} & 1231 & 799 & 9665 & 97 & 11792 \\
         \midrule
         sum & 17325 & 13249 & 182865 & 26194 & 239633 \\
         percent & 7.23\% & 5.53\% & 76.31\% & 10.93\% & 100\% \\
         \bottomrule
    \end{tabular}}
    \caption{dataset race distribution}
    \label{tab:dataset race distribution}
\end{table}

\section{Proposed Method}
\begin{figure*}[t]
\centering
\includegraphics[scale=0.35]{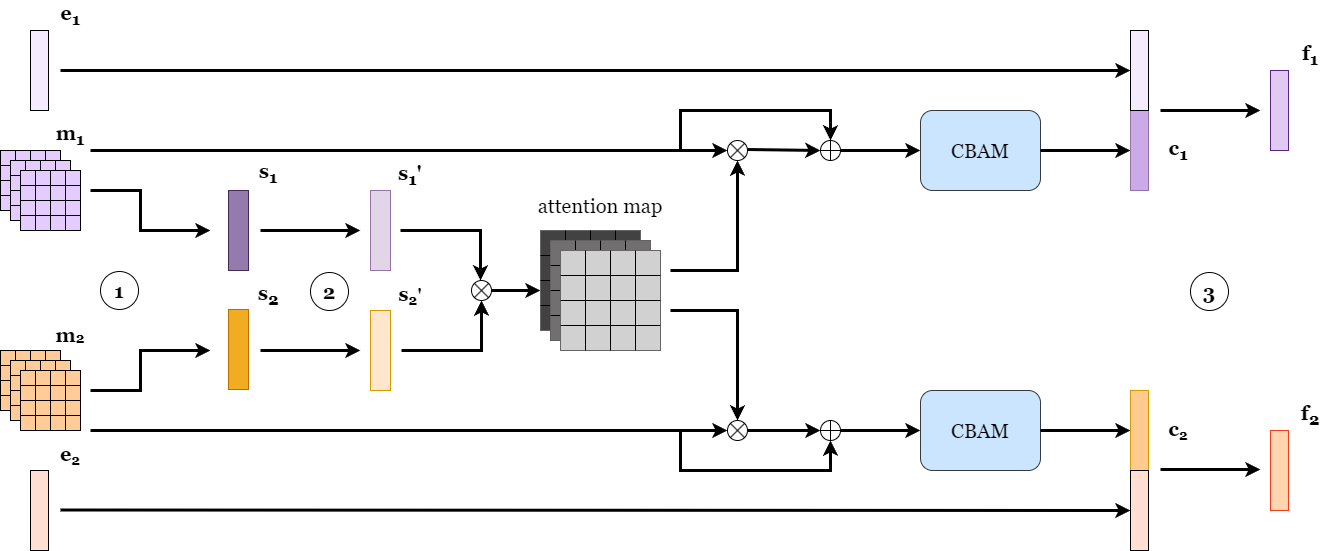}
\caption{Overview of the proposed attention module. By using the feature maps computed from the two face images to construct an attention map and employing CBAM to extract feature vectors, we concatenate the outputs with the feature embeddings to create the final output features. \textcircled 1: average pooling. \textcircled 2: 1x1 Conv with ReLU. \textcircled 3: 2 layers of 1x1 Conv with ReLU}
\label{fig:attention module}
\end{figure*}
The core idea of the proposed approach is to mitigate racial bias while improving accuracy in kinship verification task. In this section, we first introduce our model structure in detail and present why it can improve accuracy. Secondly, we analyze our fair contrastive loss function. Finally, we explain why the methods are effective for fairness and explore the reasons why they benefit racial fairness.

\subsection{Model Structure}
The proposed framework is illustrated in \autoref{fig:model structure}. For input images $(I_1, I_2) \in \mathbb{R}^{H\times W\times 3}$, we have middle layer feature maps $(m_1, m_2) \in \mathbb{R}^{H\times W\times C}$ and the outputs of backbone feature embeddings $(e_1, e_2) \in \mathbb{R}^C$. The model has a multi-task structure, the upper row is in charge of kinship verification task while the lower row predicts the race of input images.

In the upper branch, we pass feature maps into attention module to obtain representative features, and then pass them into debias layer to generate a debias term in our loss function. For our attention module in \autoref{fig:attention module}, we refer to \cite{su2023kinship} and modify their structure to reach better accuracy. We first use average pooling to obtain channel features $(s_1, s_2)$ from feature map $(m_1, m_2)$ and then use 1x1 convolution layer to get feature vectors $(s_1', s_2')$ to create attention map rather than only using one 1x1 convolution layer. This modification can not only improve the computational efficiency for downstream tasks but also allow for the interaction of different channels within a layer. Secondly, we make use of the effective attention impact of CBAM\cite{woo2018cbam}. In the original method\cite{su2023kinship}, they use two layers of 1x1 convolution to obtain feature vectors $(c_1, c_2)$, however, we use CBAM\cite{woo2018cbam} that contains channel and spatial attention, which focuses on the most relevant features of an input and generates more sophisticated features. 

Note that we also apply CBAM\cite{woo2018cbam} to our backbone in the last three layers$(\mathbb{R}^{7\times 7\times 512})$ of ResNet101\cite{he2015deep} to enhance the output feature maps and vector. Behind attention module, we pass the concatenation of two feature vectors $(e_1 \cdot c_1, e_2 \cdot c_2)$ to debias layer which is a simple linear layer for generating the debias term in loss function.

In the lower branch, we use feature vectors to predict the race. By using two layers of MLP as race classifier, we calculate cross entropy loss between the output and race ground truth. To alleviate bias, the gradient of the race classifier is reversed to remove racial information in feature vectors and decrease standard deviation. However, if the gradient is not reversed, the model can greatly improve kinship verification accuracy through multi-task learning.

\subsection{Loss Function}\label{ssec:loss function}
Supervised contrastive loss\cite{khosla2021supervised}  learns a similarity metric between samples in different classes, pulling samples from the same class closer and pushing samples from different classes apart. It outperforms other loss functions in high-dimensional feature spaces and focuses on hard samples, leading to higher accuracy. Following prior work\cite{9666944} in kinship verification task, supervised contrastive loss can be formulated as
\begin{equation}
L=\frac{1}{2n}\sum^n_{i=1}[L_c(x_i, y_i)+L_c(y_i, x_i)]
\end{equation}
where
\begin{equation}
L_c(x_i, y_i)=-log\frac{e^{s(x_i, y_i)/\tau}}{\sum^n_{j=1}[e^{s(x_i, x_j)/\tau}+e^{s(x_i, y_j)/\tau}]}
\end{equation}
where $s(x, y)$ is defined as the cosine similarity between $x$ and $y$, and $(x_i, y_i)$ are positive pairs while $(x_i, x_j), (x_i, y_j)(j \neq i)$ are negative pairs. 

\cite{wang2022mixfairface} proposed a loss function aiming to reduce identity bias in face recognition, rather than demographic bias. They pass two identities'($x_i, x_j$) feature maps into a debias layer which is a fully-connected layer to extract final feature maps$(f_i, f_j)$, and then use them to calculate the estimated bias between two identities:
\begin{equation}\label{eq:mixfairface}
cos(M(f_m), M(f_i))^2 - cos(M(f_m), M(f_j))^2 = \epsilon
\end{equation}
where $f_m=\frac{1}{2}(f_i+f_j)$ and $M(.)$ is the debias layer, and they claim that if $\epsilon > 0$ means $i$ has large biases than $j$, and vice versa. They combine the estimated bias with the cosine value and the margin in CosFace\cite{wang2018cosface}, which improves the fairness in face recognition models.

Sparked by the previous works, we propose fairness-aware contrastive loss function: 
\begin{equation}
\begin{gathered}
L_{fairness}= \\
-\log\frac{e^{( \cos(x_i, y_i) - b_{i})/\tau}}{\sum^N_{j\neq i}[e^{ \cos(x_i, x_j)/\tau} + e^{ \cos(x_i, y_j)/\tau}] + e^{( \cos(x_i, y_i) - b_{i})/\tau}}
\end{gathered}
\end{equation}

where $b_{i}$ is averaging $\epsilon$ in \autoref{eq:mixfairface} representing the bias between images $I_i$ and other images in the same batch. Following the gradient analysis in \cite{wang2021understanding}, if we minus cosine value with bias term $b_{i}$, both of the gradients of positive and negative pairs would become larger, which is similar to the margin penalty in CosFace\cite{wang2018cosface}. However, the margin is fixed in CosFace\cite{wang2018cosface}, while $b_{i}$ changes with identity pairs. After subtracting the bias term in loss function, the greater gradients help each pair balances its unfairness situation, which can make the compactness degree of every race as consistent as possible. Furthermore, the temperature $\tau$ in the original contrastive loss can further tackle with the hard samples. Combining temperature $\tau$ with the debias term, we take into account both accuracy and fairness.

In the race classifier branch, we use cross entropy(CE) loss to train the race classifier: 
\begin{equation}
L_{race}=-\sum_{i=1}^n t_i log(p_i)
\end{equation}
where $n$ is the total number of races in the training dataset, $t_i$ is the ground truth label, and $p_i$ is the softmax probability for the $i^{th}$ class. Combining these two loss functions, we have our total loss function given by
\begin{equation}
L_{total}=L_{fairness} + L_{race}
\end{equation}

\subsection{Gradients of Fair Contrastive Loss Function}
\label{ssec: gradients}
Wang and Liu\cite{wang2021understanding} analyzed the gradients with respect to positive samples $(x_i, x_i)$ and different negative samples $(x_i, x_j) (j \neq i)$: 
\begin{equation}\label{eq:gradient}
\frac{\partial L(x_i)}{\partial cos(x_i, x_i)}=-\frac{1}{\tau} \sum_{k \neq i} P_{i, k}, \hspace{1em} \frac{\partial L(x_i)}{\partial cos(x_i, x_j)}=\frac{1}{\tau} P_{i, j}
\end{equation}
where $P_{i, j}$ is the probability of $x_i$ and $x_j$ being recognized
as positive pair: 
\begin{equation}\label{eq:probability}
P_{i,j}= \frac{e^{(cos(x_i, x_j))/\tau}}{\sum_{k \neq i}e^{(cos(x_i, x_k))/\tau} + e^{(cos(x_i, x_i))/\tau}}
\end{equation}
In our case, we subtract a bias term in positive pair, which makes $P_{i, j}$ as:
\begin{equation}\label{eq:probability with bias}
P_{i,j}= \frac{e^{(cos(x_i, x_j))/\tau}}{\sum_{k \neq i}e^{(cos(x_i, x_k))/\tau} + e^{(cos(x_i, x_i)-b_i)/\tau}}
\end{equation}
When bias $b_i$ is positive, $P_{i, j}$ becomes larger, which makes the gradients of positive pair and negative pair in \autoref{eq:gradient} larger, and vice versa.

We have analyzed the average bias value during training, and it is positive. This means that both the gradients of our positive and negative pairs become larger, which makes the training more efficient and helps it converge faster.

\subsection{Fairness Mechanism}
This work employs two methods for improving fairness: adversarial learning and fair loss function. We use a race classifier in adversarial learning to remove racial information from feature vectors, which decreases standard deviation. However, due to the small dataset size compared to popular benchmark datasets and the difficulty to train a robust network using adversarial training, our results are not as outstanding as the previous works. As a result, we further modify the contrastive loss function to a fairness-aware loss function. As described in \autoref{ssec:loss function}, our loss function can eliminate bias in every two identities which indirectly improves the fairness in every race. 
Moreover, these two methods can decrease standard deviation independently which is shown in the next section, and the standard deviation becomes smaller while the accuracy is improved when we combine them together.

\section{Experiment}
\label{sec:experiment}
\subsection{Experimental Setting}
\textbf{Dataset.} \hspace{0.2cm}In this study, we employ KinRace dataset for training, validation, and testing. The dataset was divided into three sets with no overlapping families, and the ratio of the four races in each set was similar to that in \autoref{tab:dataset race distribution}. We choose four kinship relationship in our dataset: Father-Son(FS), Father-Daughter(FD), Mother-Son(MS), Mother-Daughter(MD). Face images were resized to 112 x 112 using MTCNN\cite{Zhang_2016} for face detection and alignment during preprocessing.

\noindent\textbf{Implementation Details.} We adopt ResNet101\cite{he2015deep} with ArcFace\cite{wang2022mixfairface} as the pre-trained backbone network. The sizes of feature map and feature vector  are $\mathbb{R}^{7\times 7\times 512}$ and  $\mathbb{R}^{512}$, respectively. We set the temperature parameter $\tau$ to 0.08 based on the analysis in \cite{9666944}. The models are trained with SGD algorithm, with momentum 0.9 and weight decay 1e−4. We train the networks for 10 epochs with 60000 iterations and set batch size to 25. We train a baseline\cite{9666944} which won the first place in RFIW 2021\cite{robinson20215th} to compare with our method. In the experiment below, if we mention \textbf{adversarial}, it means we reverse the gradient of race classification like the red line in the indication in \autoref{fig:model structure}. If we mention \textbf{multi-task}, it means we do not reverse the gradient of race classification, instead we just train the model normally with the green line in the indication in \autoref{fig:model structure}.

\subsection{Ablation Study}
\textbf{Effect of improving accuracy.} We conduct experiments at three techniques that improves accuracy in our methods and compare them with the baseline\cite{9666944}. The result is shown in \autoref{tab:ablation study acc}. By using attention mechanism and CBAM to extract informative features and incorporating race classification branch, our model can learn features containing racial information, leading to a nearly 7\% improvement in accuracy over the baseline. Additionally, the debias layer helps to rectify bias and make the compactness degree more uniform, resulting in a slight accuracy improvement and reduced standard deviation.


\begin{table}[h]
    \centering
    \scalebox{0.57}{
    \begin{tabular}{c|cccc|c|c}
         \toprule
         \textbf{Methods\%} & \textbf{African} & \textbf{Asian} & \textbf{Caucasian} & \textbf{Indian} & \textbf{Avg} & \textbf{Std} \\
         \midrule
         baseline & 82.18 & 83.71 & 78.00 & 80.70 & 79.08 & 2.43 \\
         KFC(attention) & 84.31 & 85.99 & 81.59 & 79.28 & 81.74 & 2.96 \\
         KFC(attention+multi-task) & 85.03 & 84.39 & 87.01 & 78.95 & 85.82 & 3.45 \\
         KFC(attention+multi-task+debias layer) & 85.16 & 84.94 & 86.67 & 81.32 & \textbf{85.86} & \textbf{2.27} \\
         \bottomrule
    \end{tabular}}
    \caption{Ablation study on accuracy.}
    \label{tab:ablation study acc}
\end{table}

\begin{table}[h]
    \centering
    \scalebox{0.57}{
    \begin{tabular}{c|cccc|c|c}
         \toprule
         \textbf{Methods\%} & \textbf{African} & \textbf{Asian} & \textbf{Caucasian} & \textbf{Indian} & \textbf{Avg} & \textbf{Std} \\
         \midrule
         baseline & 82.18 & 83.71 & 78.00 & 80.70 & 79.08 & 2.43 \\
         KFC(adversarial) & 78.45 & 81.51 & 79.11 & 76.36 & 78.88 & 2.10 \\
         KFC(debias layer) & 80.35 & 80.62 & 78.67 & 77.35 & 78.74 & 1.53 \\
         KFC(adversarial+debias layer)(acc best) & 81.61 & 82.68 & 82.90 & 82.37 & \textbf{82.74} & 0.56 \\
         KFC(adversarial+debias layer)(std best) & 81.28 & 81.29 & 80.83 & 80.80 & 80.88 & \textbf{0.27} \\
         \bottomrule
    \end{tabular}}
    \caption{Ablation study on standard deviation. \textit{acc best} refers to the epoch with the highest accuracy on validation dataset, and \textit{std best} refers to the epoch with the lowest standard deviation on validation dataset.}
    \label{tab:ablation study std}
\end{table}

\noindent\textbf{Effect of improving fairness.}
By reversing the gradient of race classification branch, the feature vector contains less racial information, which reduces the standard deviation slightly but also harms the accuracy. After adopting debias layer in our model, the debias term in loss function evidently mitigates the bias between identities, however, the accuracy goes down as well. Since these two methods can effectively improve fairness, we merge them as our final proposed method. Remarkably, the standard deviation greatly diminishes while boosting the accuracy. Firstly, the feature vector excludes racial information, which benefits from adversarial learning. Secondly, the debias layer becomes more robust because it can generate debias term depending on the most essential facial features while racial traits are removed. \autoref{tab:ablation study std} show that our strategy enhances fairness while maintaining accuracy.

\begin{table}[h]
    \centering
    \scalebox{0.72}{
    \begin{tabular}{c|cccc|c|c}
         \toprule
         \textbf{Method} & \textbf{African} & \textbf{Asian} & \textbf{Caucasian} & \textbf{Indian} & \textbf{Avg} & \textbf{Std} \\
         \midrule
         Vuvko\cite{Shadrikov_2020} & 71.13 & 73.32 & 72.61 & 76.19 & 72.96 & 3.40 \\
         Ustc-nelslip\cite{yu2020deep} & 76.05 & 77.07 & 75.54 & 63.98 & 74.33 & 6.15 \\
         TeamCNU\cite{9666944} & 82.18 & 83.71 & 78.00 & 80.70 & 79.08 & 2.43 \\
         KFC(multi-task) & 85.16 & 84.94 & 86.67 & 81.32 & \textbf{85.86} & 2.27 \\
         KFC(adversarial) & 81.28 & 81.29 & 80.83 & 80.80 & 80.88 & \textbf{0.27} \\
         \bottomrule
    \end{tabular}}
    \caption{Comparisons with SOTA methods on KinRace dataset}
    \label{tab:SOTA comparison on KinRace}
\end{table}

\begin{table}
    \centering
    \scalebox{0.8}{
    \begin{tabular}{c|cc|c|c}
         \toprule
         \textbf{Method} & \textbf{C \& YP} & \textbf{C \& OP} & \textbf{Avg} & \textbf{Std} \\
         \midrule
         LPQ\_ML\cite{inproceedings2} & - & - & 73.25 & - \\
         StatBIF-SIWEDA-WCCN\cite{article} & 75.71 & 76.92 & 76.32 & - \\
         FAML\cite{10.1007/s11042-022-12032-w} & 78.30 & 75.00 & 76.54 & - \\
         B$\textrm{C}^2$DA\cite{MUKHERJEE2022116829} & 83.28 & \textbf{82.69} & 83.30 & - \\
         KFC(multi-task) & \textbf{87.24} & 81.00 & \textbf{84.12} & 1.36 \\
         KFC(adversarial) & 82.71 & 77.75 & 80.23 & \textbf{0.06} \\
         \bottomrule
    \end{tabular}}
    \caption{Comparisons with SOTA methods on UB KinFace dataset. This dataset provides child, young and old parent. C \& YP means child and young parent, and C \& OP means child and old parent.}
    \label{tab:SOTA comparison on UB KinFace}
\end{table}

\begin{table}
    \centering
    \scalebox{0.8}{
    \begin{tabular}{c|cccc|c}
         \toprule
         \textbf{Method} & \textbf{FD} & \textbf{MD} & \textbf{FS} & \textbf{MS} & \textbf{Std} \\
         \midrule
         Vuvko\cite{Shadrikov_2020} & 75.00 & 78.00 & 81.00 & 74.00 & - \\
         Ustc-nelslip\cite{yu2020deep} & 76.00 & 75.00 & 82.00 & 75.00 & - \\
         TeamCNU\cite{9666944} & 75.00 & 80.00 & 82.00 & 77.0 & - \\
         FaCoRNet(ArcFace)\cite{su2023kinship} & 77.30 & 80.40 & 82.60 & 78.80 & - \\
         FaCoRNet(AdaFace)\cite{su2023kinship} & \textbf{79.50} & 81.80 & \textbf{84.80} & \textbf{80.20} & - \\
         KFC(multi-task) & 79.05 & \textbf{83.61} & 84.63 & 78.25 & 7.81 \\
         KFC(adversarial) & 78.81 & 82.56 & 81.69 & 77.43 & \textbf{5.57} \\
         \bottomrule
    \end{tabular}}
    \caption{Comparisons with SOTA methods on FIW dataset. Since we only focus on four kinship relations, we compare these four types with SOTA methods.}
    \label{tab:SOTA comparison on FIW}
\end{table}

\subsection{Comparisons with SOTA methods}
\label{ssec:compare with SOTA}
\begin{figure}
\centering
\includegraphics[scale=0.53]{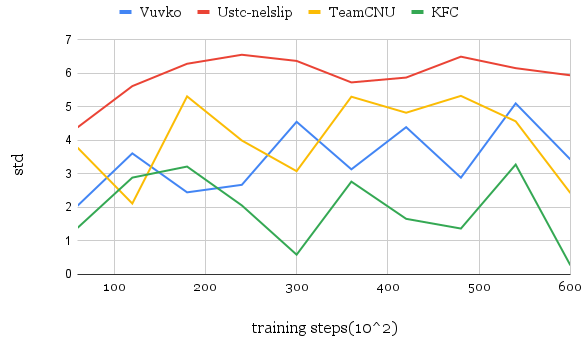}
\caption{Comparison on standard deviation variations for different methods during training.}
\label{fig:std chart}
\end{figure}

In this section, we compare our method with three other works\cite{Shadrikov_2020}\cite{yu2020deep}\cite{9666944} that performed well in the RFIW challenge\cite{9320293}\cite{robinson20215th}. We re-implement these methods using their training settings and evaluate them on our KinRace dataset. Our proposed method outperforms these methods in both accuracy and standard deviation, as shown in \autoref{tab:SOTA comparison on KinRace}. Our multi-task method achieves almost 13\% gain for average accuracy, while our adversarial method reduce the standard deviation to 0.27 which tremendously improves fairness. In \autoref{fig:std chart}, we calculate the standard deviation on our KinRace testing set every 10000 iteration on SOTA methods and our proposed KFC. It is obviously that the experimental result well demonstrates fairness of the proposed KFC method, which effectively decreases the standard deviation to a very low level. 

To evaluate the generalization of our method, we evaluate our method on other datasets: UB KinFace\cite{Buffalo_TMM_Kinship}\cite{Siyu_IJCAI11_Kinship}\cite{Ming_CVPR11_Genealogical} and FIW\cite{robinson2016families}. The results of UB KinFace\cite{Buffalo_TMM_Kinship}\cite{Siyu_IJCAI11_Kinship}\cite{Ming_CVPR11_Genealogical} is shown in \autoref{tab:SOTA comparison on UB KinFace}, our multi-task version outperforms previous works, while the adversarial version diminishes the standard deviation significantly. We follow the standard protocol\cite{Buffalo_TMM_Kinship}\cite{Siyu_IJCAI11_Kinship}\cite{Ming_CVPR11_Genealogical} with five-fold cross-validation. In \autoref{tab:SOTA comparison on FIW}, we compare 4 kinship relations with SOTA methods. Although our accuracy may not be the highest, our results are competitive compared to FaCoRNet(AdaFace)\cite{su2023kinship}, and our backbone is ArcFace, which actually makes our results better than FaCoRNet(ArcFace)\cite{su2023kinship}. Additionally, our adversarial version reduces the standard deviation compared to the multi-task version, as shown in \autoref{tab:SOTA comparison on UB KinFace} and \autoref{tab:SOTA comparison on FIW}, and more details will be provided in supplementary material.
In these experiments, we demonstrate that our proposed KFC can perform well on datasets of different sizes, outperforming state-of-the-art methods and effectively decreasing standard deviation.

\subsection{Visualization and Analysis on Fairness}
\label{sec: visualization and analysis}
\begin{figure}
    \centering
    \subfigure[baseline]{\includegraphics[width=0.215\textwidth]{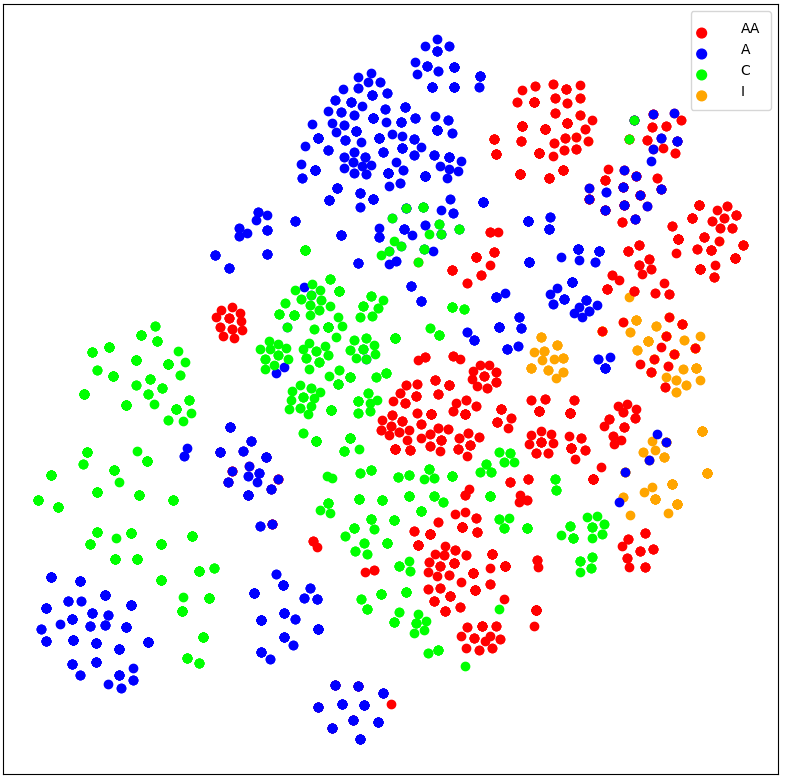}}
    \qquad
    \subfigure[KFC]{\includegraphics[width=0.215\textwidth]{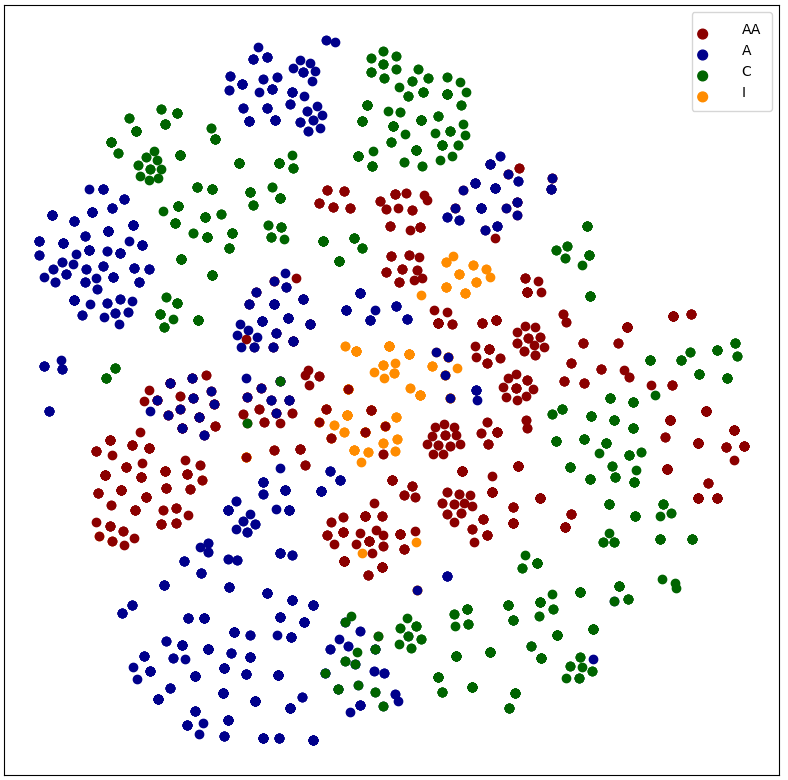}}
    \caption{t-SNE visualizations. We randomly pick 400 pairs per race from KinRace dataset. AA for African, A for Asian, C for Caucasian, and I for Indian}
    \label{fig:t-SNE}
\end{figure}

To prove our proposed method can genuinely improve fairness, we analyze intra-class and inter-class angle between baseline\cite{9666944} and our KFC. In \autoref{tab:intra-class and inter class angle comparison}, our KFC balances the intra-class and inter-class angle between four races, which makes the compactness degree comparable, and thus markedly improves fairness. As shown in \autoref{fig:t-SNE}, the feature embeddings for each race are distributed more evenly than the baseline\cite{9666944}. In the baseline graph, Asian and Caucasian groups are mainly on the left while African and Indian groups are mainly on the right. Our adversarial learning approach removes clear boundaries between the four races, providing a more fair solution for kinship verification, supported by our experimental results and feature space analysis.

\begin{table}[h]
    \centering
    \scalebox{0.8}{
    \begin{tabular}{c|c|cccc|c}
         \toprule
         \multicolumn{2}{c|}{\textbf{Methods}} & \textbf{African} & \textbf{Asian} & \textbf{Caucasian} & \textbf{Indian} & \textbf{Std} \\
         \hline
         \multirow{2}{*}{baseline\cite{9666944}} & intra & 17.21 & 18.32 & 15.36 & 10.08 & 3.68 \\
         & inter & 46.85 & 52.34 & 40.85 & 44.20 & 4.85 \\
         \hline
         \multirow{2}{*}{ours} & intra & 12.21 & 16.59 & 15.65 & 11.36 & \textbf{2.47} \\
         & inter & 41.35 & 49.93 & 41.04 & 42.80 & \textbf{4.17} \\
         \bottomrule
    \end{tabular}}
    \caption{Intra-class and inter-class angle comparison. We randomly select 20 families per race from KinRace dataset.}
    \label{tab:intra-class and inter class angle comparison}
\end{table}

\section{Conclusion}
This paper focuses on mitigating the issue of racial bias in kinship verification by using fair contrastive loss function and adversarial learning while improving the accuracy by utilizing attention module and multi-task learning. We also provide a kinship dataset KinFace with racial labels to broaden kinship research area to a more diversified scenario. Our experimental results suggest that our combination of model structure, gradient design, and fair loss function can significantly improve the racial fairness for kinship verification while automatically adjust intra- and inter-class angles in feature space, thus leading to standard deviation reduction and accuracy boost simultaneously.

{\small
\bibliographystyle{ieee_fullname}
\bibliography{egbib}

\begin{thebibliography}{10}\itemsep=-1pt

\bibitem{inproceedingsLBP}
Timo Ahonen, Abdenour Hadid, and Matti Pietikäinen.
\newblock Face recognition with local binary patterns.
\newblock volume 3021, pages 469--481, 05 2004.

\bibitem{10.5555/2987189.2987282}
Jane Bromley, Isabelle Guyon, Yann LeCun, Eduard S\"{a}ckinger, and Roopak Shah.
\newblock Signature verification using a "siamese" time delay neural network.
\newblock In {\em Proceedings of the 6th International Conference on Neural Information Processing Systems}, NIPS'93, page 737–744, San Francisco, CA, USA, 1993. Morgan Kaufmann Publishers Inc.

\bibitem{inproceedings2}
Abdelhakim Chergui, Salim Ouchtati, Hichem Telli, Fares Bougourzi, and Salah~Eddine Bekhouche.
\newblock Lpq and ldp descriptors with ml representation for kinship verification.
\newblock 04 2018.

\bibitem{Chicco2021}
Davide Chicco.
\newblock {\em Siamese Neural Networks: An Overview}, pages 73--94.
\newblock Springer US, New York, NY, 2021.

\bibitem{1467314}
S. Chopra, R. Hadsell, and Y. LeCun.
\newblock Learning a similarity metric discriminatively, with application to face verification.
\newblock In {\em 2005 IEEE Computer Society Conference on Computer Vision and Pattern Recognition (CVPR'05)}, volume~1, pages 539--546 vol. 1, 2005.

\bibitem{8019326}
Lvye Cui and Bo Ma.
\newblock Adaptive feature selection for kinship verification.
\newblock In {\em 2017 IEEE International Conference on Multimedia and Expo (ICME)}, pages 751--756, 2017.

\bibitem{Deng_2022}
Jiankang Deng, Jia Guo, Jing Yang, Niannan Xue, Irene Kotsia, and Stefanos Zafeiriou.
\newblock {ArcFace}: Additive angular margin loss for deep face recognition.
\newblock {\em {IEEE} Transactions on Pattern Analysis and Machine Intelligence}, 44(10):5962--5979, oct 2022.

\bibitem{6738614}
Ruogu Fang, Andrew~C. Gallagher, Tsuhan Chen, and Alexander Loui.
\newblock Kinship classification by modeling facial feature heredity.
\newblock In {\em 2013 IEEE International Conference on Image Processing}, pages 2983--2987, 2013.

\bibitem{5652590}
Ruogu Fang, Kevin~D. Tang, Noah Snavely, and Tsuhan Chen.
\newblock Towards computational models of kinship verification.
\newblock In {\em 2010 IEEE International Conference on Image Processing}, pages 1577--1580, 2010.

\bibitem{gong2020jointly}
Sixue Gong, Xiaoming Liu, and Anil~K. Jain.
\newblock Jointly de-biasing face recognition and demographic attribute estimation, 2020.

\bibitem{gong2020mitigating}
Sixue Gong, Xiaoming Liu, and Anil~K. Jain.
\newblock Mitigating face recognition bias via group adaptive classifier, 2020.

\bibitem{2018}
The Guardian.
\newblock Amazon ditched ai recruiting tool that favored men for technical jobs, 2018.
\newblock October 11, 2018.

\bibitem{guo2016msceleb1m}
Yandong Guo, Lei Zhang, Yuxiao Hu, Xiaodong He, and Jianfeng Gao.
\newblock Ms-celeb-1m: A dataset and benchmark for large-scale face recognition, 2016.

\bibitem{he2015deep}
Kaiming He, Xiangyu Zhang, Shaoqing Ren, and Jian Sun.
\newblock Deep residual learning for image recognition, 2015.

\bibitem{2016}
Lauren~Kirchner Jeff~Larson, Surya~Mattu and Julia Angwin.
\newblock How we analyzed the compas recidivism algorithm, 2016.
\newblock May 23, 2016.

\bibitem{khosla2021supervised}
Prannay Khosla, Piotr Teterwak, Chen Wang, Aaron Sarna, Yonglong Tian, Phillip Isola, Aaron Maschinot, Ce Liu, and Dilip Krishnan.
\newblock Supervised contrastive learning, 2021.

\bibitem{article}
Oualid Laiadi, Abdelmalik Ouamane, · Benakcha, Abdelmalik taleb ahmed, and Abdenour Hadid.
\newblock A weighted exponential discriminant analysis through side-information for face and kinship verification using statistical binarized image features.
\newblock {\em International Journal of Machine Learning and Cybernetics}, 12, 01 2021.

\bibitem{6247978}
Jiwen Lu, Junlin Hu, Xiuzhuang Zhou, Yuanyuan Shang, Yap-Peng Tan, and Gang Wang.
\newblock Neighborhood repulsed metric learning for kinship verification.
\newblock In {\em 2012 IEEE Conference on Computer Vision and Pattern Recognition}, pages 2594--2601, 2012.

\bibitem{Ming_CVPR11_Genealogical}
S.~Xia M.~Shao and Y. Fu.
\newblock Genealogical face recognition based on ub kinface database.
\newblock In {\em Proc. IEEE CVPR Workshop on Biometrics (BIOM)}, 2011.

\bibitem{MUKHERJEE2022116829}
Moumita Mukherjee and Toshanlal Meenpal.
\newblock Binary cross coupled discriminant analysis for visual kinship verification.
\newblock {\em Signal Processing: Image Communication}, 108:116829, 2022.

\bibitem{inproceedings}
Abhilash Nandy and Shanka Mondal.
\newblock Kinship verification using deep siamese convolutional neural network.
\newblock pages 1--5, 05 2019.

\bibitem{park2022fair}
Sungho Park, Jewook Lee, Pilhyeon Lee, Sunhee Hwang, Dohyung Kim, and Hyeran Byun.
\newblock Fair contrastive learning for facial attribute classification, 2022.

\bibitem{10.1007/s11042-022-12032-w}
Xiaoqian Qin, Dakun Liu, and Dong Wang.
\newblock A novel factor analysis-based metric learning method for kinship verification.
\newblock {\em Multimedia Tools Appl.}, 81(8):11049–11070, mar 2022.

\bibitem{raff2018gradient}
Edward Raff and Jared Sylvester.
\newblock Gradient reversal against discrimination, 2018.

\bibitem{robinson20215th}
Joseph~P. Robinson, Can Qin, Ming Shao, Matthew~A. Turk, Rama Chellappa, and Yun Fu.
\newblock The 5th recognizing families in the wild data challenge: Predicting kinship from faces, 2021.

\bibitem{robinson2016families}
Joseph~P. Robinson, Ming Shao, Yue Wu, and Yun Fu.
\newblock Families in the wild (fiw): Large-scale kinship image database and benchmarks.
\newblock In {\em Proceedings of the 2016 ACM on Multimedia Conference}, pages 242--246. ACM, 2016.

\bibitem{9320293}
Joseph~P. Robinson, Yu Yin, Zaid Khan, Ming Shao, Siyu Xia, Michael Stopa, Samson Timoner, Matthew~A. Turk, Rama Chellappa, and Yun Fu.
\newblock Recognizing families in the wild (rfiw): The 4th edition.
\newblock In {\em 2020 15th IEEE International Conference on Automatic Face and Gesture Recognition (FG 2020)}, pages 857--862, 2020.

\bibitem{Buffalo_TMM_Kinship}
J.~Luo S.~Xia, M.~Shao and Y. Fu.
\newblock Understanding kin relationships in a photo.
\newblock {\em IEEE Transactions on Multimedia}, 14(4):1046--1056, 2012.

\bibitem{Siyu_IJCAI11_Kinship}
M.~Shao S.~Xia and Y. Fu.
\newblock Kinship verification through transfer learning.
\newblock In {\em Proc. International Joint Conferences on Artificial Intelligence (IJCAI)}, pages 2539--2544, 2011.

\bibitem{Shadrikov_2020}
Andrei Shadrikov.
\newblock Achieving better kinship recognition through better baseline.
\newblock In {\em 2020 15th {IEEE} International Conference on Automatic Face and Gesture Recognition ({FG} 2020)}. {IEEE}, nov 2020.

\bibitem{su2023kinship}
Weng-Tai Su, Min-Hung Chen, Chien-Yi Wang, Shang-Hong Lai, and Trista Pei-Chun Chen.
\newblock Kinship representation learning with face componential relation.
\newblock In {\em Proceedings of the IEEE/CVF International Conference on Computer Vision (ICCV) Workshops}, 2023.

\bibitem{6909616}
Yaniv Taigman, Ming Yang, Marc'Aurelio Ranzato, and Lior Wolf.
\newblock Deepface: Closing the gap to human-level performance in face verification.
\newblock In {\em 2014 IEEE Conference on Computer Vision and Pattern Recognition}, pages 1701--1708, 2014.

\bibitem{2019}
Starre Vartan.
\newblock Racial bias found in a major health care risk algorithm, 2019.
\newblock October 24, 2019.

\bibitem{wang2021understanding}
Feng Wang and Huaping Liu.
\newblock Understanding the behaviour of contrastive loss, 2021.

\bibitem{wang2022mixfairface}
Fu-En Wang, Chien-Yi Wang, Min Sun, and Shang-Hong Lai.
\newblock Mixfairface: Towards ultimate fairness via mixfair adapter in face recognition, 2022.

\bibitem{wang2018cosface}
Hao Wang, Yitong Wang, Zheng Zhou, Xing Ji, Dihong Gong, Jingchao Zhou, Zhifeng Li, and Wei Liu.
\newblock Cosface: Large margin cosine loss for deep face recognition, 2018.

\bibitem{Wang_2019_ICCV}
Mei Wang, Weihong Deng, Jiani Hu, Xunqiang Tao, and Yaohai Huang.
\newblock Racial faces in the wild: Reducing racial bias by information maximization adaptation network.
\newblock In {\em The IEEE International Conference on Computer Vision (ICCV)}, October 2019.

\bibitem{9512390}
Mei Wang, Yaobin Zhang, and Weihong Deng.
\newblock Meta balanced network for fair face recognition.
\newblock {\em IEEE Transactions on Pattern Analysis and Machine Intelligence}, 44(11):8433--8448, 2022.

\bibitem{wang2022fairnessaware}
Zhibo Wang, Xiaowei Dong, Henry Xue, Zhifei Zhang, Weifeng Chiu, Tao Wei, and Kui Ren.
\newblock Fairness-aware adversarial perturbation towards bias mitigation for deployed deep models, 2022.

\bibitem{woo2018cbam}
Sanghyun Woo, Jongchan Park, Joon-Young Lee, and In~So Kweon.
\newblock Cbam: Convolutional block attention module, 2018.

\bibitem{10.1007/s11263-022-01605-9}
Xiaoting Wu, Xiaoyi Feng, Xiaochun Cao, Xin Xu, Dewen Hu, Miguel~Bordallo L\'{o}pez, and Li Liu.
\newblock Facial kinship verification: A comprehensive review and outlook.
\newblock {\em Int. J. Comput. Vision}, 130(6):1494–1525, jun 2022.

\bibitem{xu2021consistent}
Xingkun Xu, Yuge Huang, Pengcheng Shen, Shaoxin Li, Jilin Li, Feiyue Huang, Yong Li, and Zhen Cui.
\newblock Consistent instance false positive improves fairness in face recognition, 2021.

\bibitem{YAN2019169}
Haibin Yan and Shiwei Wang.
\newblock Learning part-aware attention networks for kinship verification.
\newblock {\em Pattern Recognition Letters}, 128:169--175, 2019.

\bibitem{yu2020deep}
Jun Yu, Mengyan Li, Xinlong Hao, and Guochen Xie.
\newblock Deep fusion siamese network for automatic kinship verification, 2020.

\bibitem{Zhang_2016}
Kaipeng Zhang, Zhanpeng Zhang, Zhifeng Li, and Yu Qiao.
\newblock Joint face detection and alignment using multitask cascaded convolutional networks.
\newblock {\em {IEEE} Signal Processing Letters}, 23(10):1499--1503, oct 2016.

\bibitem{9666944}
Ximiao Zhang, Min XU, Xiuzhuang Zhou, and Guodong Guo.
\newblock Supervised contrastive learning for facial kinship recognition.
\newblock In {\em 2021 16th IEEE International Conference on Automatic Face and Gesture Recognition (FG 2021)}, pages 01--05, 2021.

\end{thebibliography}
}

\end{document}